%% file: main.tex
\documentclass[letterpaper, conference]{ieeeconf}
\IEEEoverridecommandlockouts %

\usepackage{graphicx} 
\usepackage{amsmath, amsfonts, amssymb}

\usepackage{amsthm}
\usepackage{hyperref}
\usepackage{booktabs}
\input{tikz_preamble}
\title{\LARGE\bfseries Multi-CONDOR: Learning Multistable Motion Primitives from Demonstration}

\usepackage{verbatim}
\author{\authorblockN{Gabriele Giuseppe Di Marzo$^{1, 4,*, \dagger}$, Mariano Ram\'irez Montero$^{1,*, \dagger}$,\\ Maximilian St\"olzle$^{1,2}$, Rodrigo P\'erez-Dattari$^{1,3}$, and Cosimo Della Santina$^{1}$}
  \thanks{This work was supported by the European Union's Horizon Europe program through the EIC EMERGE project (Grant Agreement No.~101070918).}
  \thanks{*These authors contributed equally to this work.}
  \thanks{$\dagger$Corresponding author.}
  \thanks{$^{1}$Department of Cognitive Robotics, Delft University of Technology, Delft, The Netherlands. \texttt{\{m.ramirezmontero-1, c.dellasantina\}@tudelft.nl}.}
  \thanks{$^{2}$Computer Science and Artificial Intelligence Laboratory, Massachusetts Institute of Technology, Cambridge, MA, USA.
    \texttt{mstolzle@mit.edu}.}
  \thanks{$^{3}$Department of Robotics, Perception and Learning, KTH Royal Institute of Technology, Stockholm, Sweden.
    \texttt{rpd@kth.se}.}
  \thanks{$^{4}$Department of Computer, Control and Management Engineering (DIAG), Sapienza University of Rome, Italy. \texttt{gabriele.dimarzo@fisic.ai}}
  }
\newtheorem{theorem}{Theorem}

\theoremstyle{remark}
\newtheorem{remark}{Remark}

\begin{document}

\maketitle

\begin{abstract}
Dynamical-system-based motion policies provide a framework for learning reactive robot motions with stability and convergence properties.
Many manipulation tasks, however, admit several distinct goals or a continuous set of acceptable outcomes, requiring motion primitives that preserve these alternatives.
Accommodating both goal structures within a unified framework with continuous, autonomous dynamics remains an open challenge.
We propose a CONDOR-based framework, multi-CONDOR, for learning motion primitives with either multiple discrete attractors or a continuous attracting goal contour.
The approach combines imitation learning with losses that encourage agreement between the learned task dynamics and latent reference systems encoding the desired attraction structure.
We provide a Lyapunov-based analysis of the convergence properties of the learned motion dynamics.
We evaluate the framework on simulated benchmarks with discrete goals, qualitative examples with continuous goal contours, and robotic grasping experiments.
In simulation, multi-CONDOR reduces trajectory RMSE by approximately 72\% compared with separately learned single-goal motion primitives selected by a state machine.
For tasks with two and three goals, respectively, it reduces the fraction of evaluated initial states ending outside the goal tolerance from StableNODE's 0.31\% and 7.29\% to 0\% in both settings.
Experiments with a Franka Emika robot arm demonstrate successful grasping of a container by either of its two handles using motions learned from demonstrations.
Together, these results show multi-CONDOR combines motion reproduction with the flexibility to converge to alternative valid outcomes.
\end{abstract}
\begin{figure*}[hbtp!]
    \centering
    \includegraphics[width=0.8\linewidth, trim=0cm 3.5cm 0cm 3cm,clip]{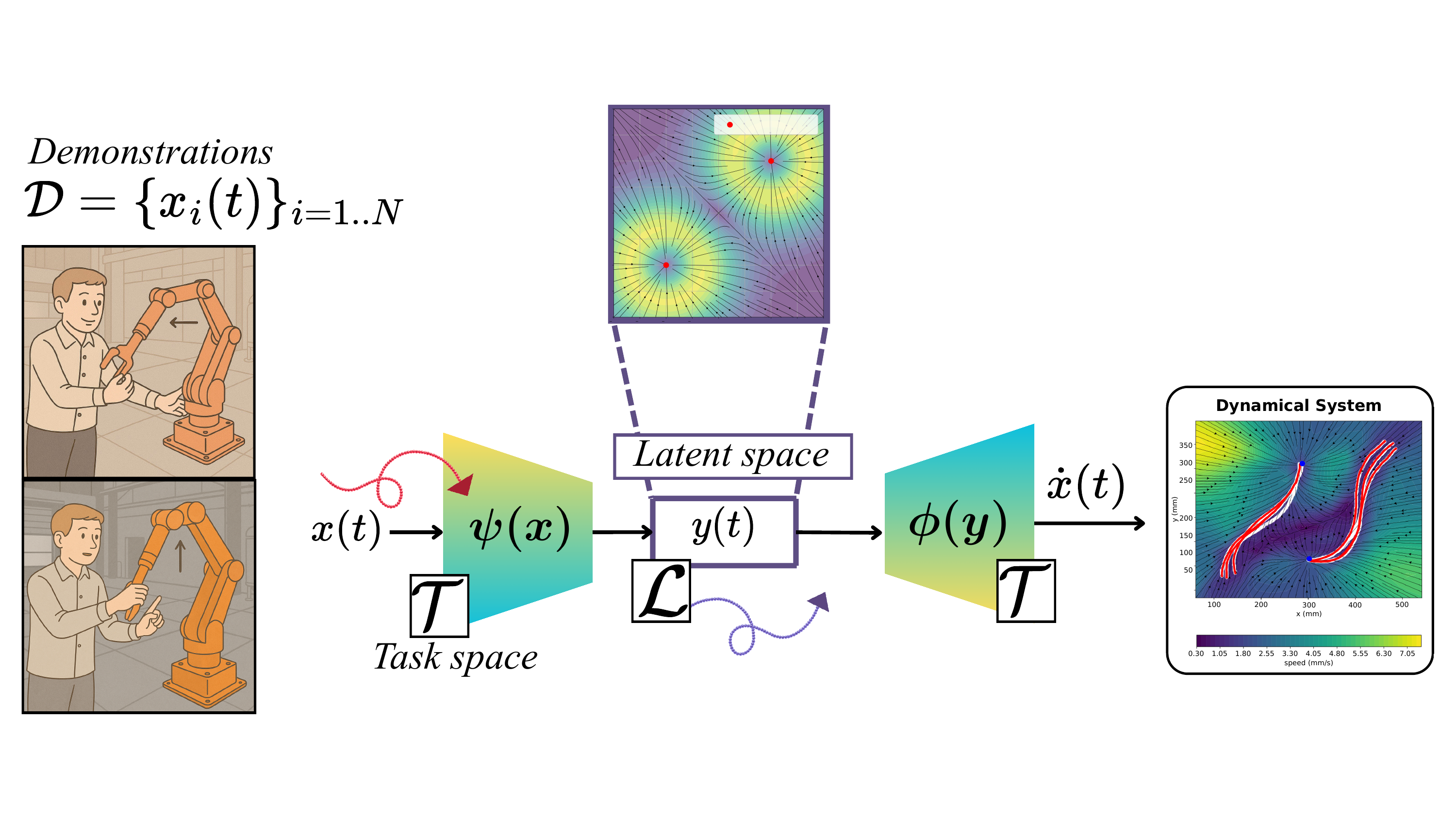}
    \caption{Overview of multi-CONDOR: an encoder $\psi$ maps task states $x(t)$ to latent states $y(t)$, and a function $\phi$ maps these to task velocities $\dot{x}(t)$. Both are jointly trained as parts of a single neural network. Training aligns the encoded task dynamics with prescribed latent reference dynamics featuring multiple attractors (shown above ``Latent space''), encouraging the learned vector field to inherit their multistability while reproducing the demonstrations~$\mathcal{D}$.}
    \label{fig:method_summary}
\end{figure*}
\vspace{-1mm}
\section{Introduction}
Imitation learning enables robots to acquire motion skills from demonstrations, providing an accessible way to adapt their behavior to new tasks.
This capability is increasingly valuable as robots enter environments where tasks vary and frequent manual programming is impractical.
General approaches, including generative methods such as diffusion policies~\cite{chi2023DiffusionPolicy}, can represent complex behaviors, but may require substantial demonstration data~\cite{wang2025EquivariantDiffusionPolicy} and do not inherently provide stability or convergence guarantees.

Dynamical systems provide an established framework for learning motion primitives with stability and convergence properties.
Dynamic movement primitives (DMPs) reproduce demonstrated motions using an internal phase variable that governs their progression~\cite{ijspeert2013DynamicalMovementPrimitives}.
If this phase advances while physical interaction slows or blocks the robot~\cite{karlsson2018DmpTemporalCoupling}, the resulting mismatch can induce unnatural catch-up motions~\cite{schaal2003MotorLearningByImitation}.
Other approaches learn autonomous vector fields that determine motion directly from the current task state, without an independent time or phase reference~\cite{khansariZadeh2011StableDynamicalSystems,rana2020EuclideanizingFlows,perezDattari2023Condor,perezDattari2024Puma}.

Standard point-to-point motion primitives typically encode convergence to a single prescribed goal~\cite{ijspeert2013DynamicalMovementPrimitives,saveriano2023DmpSurvey}.
Many manipulation tasks instead admit multimodal goal behavior: a robot may grasp a container at either of two handles.
Other tasks admit a continuum of acceptable outcomes, such as grasping at any admissible point along its rim.
Selecting one endpoint in advance discards alternatives that may be easier to reach from the current configuration or after a disturbance.
A disturbance can then redirect the motion toward another valid goal, allowing the robot to complete the task without returning to the originally selected endpoint.
Motion primitives should therefore capture both the demonstrated approach behavior and the freedom to converge to multiple isolated goals or a continuous goal set.

Initial work has addressed the discrete multistable case.
Augmented-SVM combines motion dynamics through learned partitions of the state space and modulates their vector fields to preserve the associated attraction regions~\cite{shukla2012AugmentedSvm}.
StableNODE learns latent dynamics with multiple attractors using a corrective term derived from an energy function~\cite{sochopoulos2024StableNeuralOdes}.
These approaches demonstrate motion generation toward distinct goals, but do not provide a unified framework for learning continuous, autonomous motion dynamics with both multiple discrete attractors and continuous attracting goal sets.

We propose a unified framework for learning motion primitives with continuous, autonomous dynamics and either multiple discrete attractors or a continuous attracting goal contour.
Building on CONDOR~\cite{perezDattari2023Condor}, we learn a task-space vector field together with an encoding into a latent space equipped with reference dynamics representing the desired attraction structure.
We introduce two reference systems: a Gaussian potential for multiple isolated goals and a radially attracting circle for a continuous goal contour.
We establish sufficient conditions for convergence to our desired set through Lyapunov analysis, and formulate training losses that encourage these conditions alongside an imitation objective.
The resulting framework learns motion toward alternative admissible goals directly from the current state.

We evaluate the framework on two-dimensional demonstration datasets with two and three discrete goals, measuring trajectory reproduction and convergence from initial states distributed throughout the workspace.
We further examine continuous goal contours through qualitative examples and demonstrate robotic grasping with multiple admissible grasping locations.
Together, these experiments assess the framework's ability to reproduce demonstrated motions while accommodating discrete and continuous goal sets.
%
%
%
\section{Methodology}
Given motion demonstrations in a task domain
$\mathcal{T}\subset\mathbb{R}^{N_x}$ and a set of admissible goals
$\mathcal{G}\subset\mathcal{T}$, we aim to learn an autonomous dynamical system
\begin{equation}
    \dot{x}=f_{\theta}(x),
\label{eq:task_dynamics}
\end{equation}
that reproduces the demonstrated motion while approaching a goal in
$\mathcal{G}$. We consider both multiple isolated goals and a continuous
attracting goal contour. The velocity depends only on the current state,
with no explicit dependence on time or a phase variable, so the eventual
goal is determined by the dynamics from the current state.
We build on CONDOR~\cite{perezDattari2023Condor}, which represents the task dynamics as a
neural network $f_\theta=\phi_\theta\circ\psi_\theta$. The encoder
$\psi_\theta:\mathcal{T}\to\mathbb{R}^{N_y}$ maps a task state $x$ to a latent
state $y=\psi_\theta(x)$, and
$\phi_\theta:\mathbb{R}^{N_y}\to\mathbb{R}^{N_x}$ maps this latent state to a
task velocity. CONDOR trains the encoded task trajectories to follow
prescribed latent reference dynamics $\dot y=f_{\mathcal L}(y)$ with a single
asymptotically stable equilibrium. We extend this framework by introducing
two reference systems that encode either multiple discrete attractors or a
continuous attracting contour.
Training combines an imitation objective with a stability loss comprising
trajectory matching and separation to discourage collapsed encodings.
The task dynamics inherit convergence properties of the reference system
under the conditions developed below, and our approach is summarized in \autoref{fig:method_summary}. We first introduce the stability loss
(Section~\ref{sec:stability_loss}) and the convergence conditions
(Section~\ref{sec:conditions}), then present the two reference systems and
their additional losses (Sections~\ref{sec:discrete_goals}
and~\ref{sec:continuous_goals}), and finally give the imitation loss and
complete training objectives (Section~\ref{sec:imitation_loss}).
\vspace{-0mm}
\subsection{Stability loss}
\label{sec:stability_loss}
\begin{figure}[hbtp!]
    \centering
  \resizebox{\columnwidth}{!}{%
    \input{matching_tikz}%
  }
    \caption{Matching loss for two sampled initial task states $x_{0,b}$.
    Encoded task rollouts $y^{\mathcal T}_{t,b}=\psi_\theta(x_{t,b})$ are
    compared with reference rollouts $y^{\mathcal L}_{t,b}$ initialized at
    the same latent state $\psi_\theta(x_{0,b})$.}
    \label{fig:matching}
\end{figure}
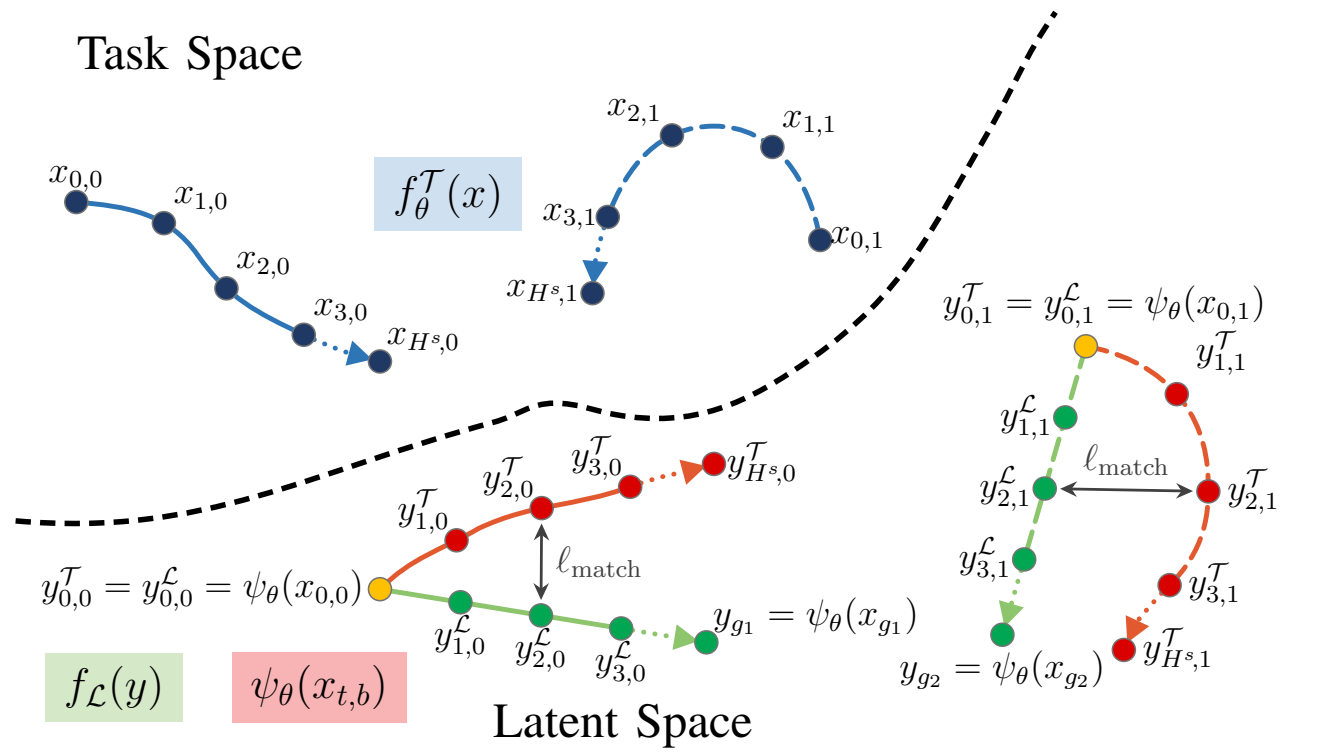

We sample $B^s$ initial states $x_{0,b}$ uniformly from $\mathcal{T}$ and roll
out the task and reference dynamics for $H^s$ steps with time step $\Delta t$.
Using $y^{\mathcal L}_{0,b}=\psi_\theta(x_{0,b})$, the rollouts are
\begin{equation}
\begin{aligned}
x_{t,b}&=x_{t-1,b}+\Delta t\,f_\theta(x_{t-1,b}),\\
y^{\mathcal L}_{t,b}&=y^{\mathcal L}_{t-1,b}+\Delta t\,f_{\mathcal L}(y^{\mathcal L}_{t-1,b}).
\end{aligned}
\label{eq:stability_rollouts}
\end{equation}
The matching loss encourages the encoded task trajectories
$y^{\mathcal T}_{t,b}=\psi_\theta(x_{t,b})$ to follow the reference dynamics
and reproduce their intended attraction behavior by penalizing differences
between the two rollouts (\autoref{fig:matching}):
\begin{equation}
\ell_{\mathrm{match}}=\frac{1}{B^sH^s}
\sum_{b=0}^{B^s-1}\sum_{t=1}^{H^s}\bigl\|y^{\mathcal L}_{t,b}-y^{\mathcal T}_{t,b}\bigr\|_2^2.
\label{eq:match_loss}
\end{equation}
The matching loss alone admits collapsed encodings: mapping every task
state to the same latent equilibrium makes it zero regardless of the task
dynamics. We therefore introduce a separation loss that encourages distinct
consecutive task states to remain distinguishable in latent space:
\begin{equation}
\ell_{\mathrm{sep}}=\frac{1}{B^sH^s}
\sum_{b=0}^{B^s-1}\sum_{t=1}^{H^s}
\max\bigl(0,\,m-\|y^{\mathcal T}_{t,b}-y^{\mathcal T}_{t-1,b}\|_2\bigr)^2,
\label{eq:sep_loss}
\end{equation}
which penalizes consecutive latent states separated by less than a margin
$m$. The separation penalty remains active near equilibria, including the
desired goals; a small margin limits its influence in these regions.
Our stability loss is
\begin{equation}
\ell_{\mathrm{stable}}=\ell_{\mathrm{match}}+\ell_{\mathrm{sep}}.
\label{eq:stability_loss}
\end{equation}
Section~\ref{sec:conditions} states sufficient conditions for convergence and
relates them to these losses.
\subsection{Conditions for multistable dynamics}
\label{sec:conditions}
\begin{figure}
    \centering
    \includegraphics[width=0.98\linewidth]{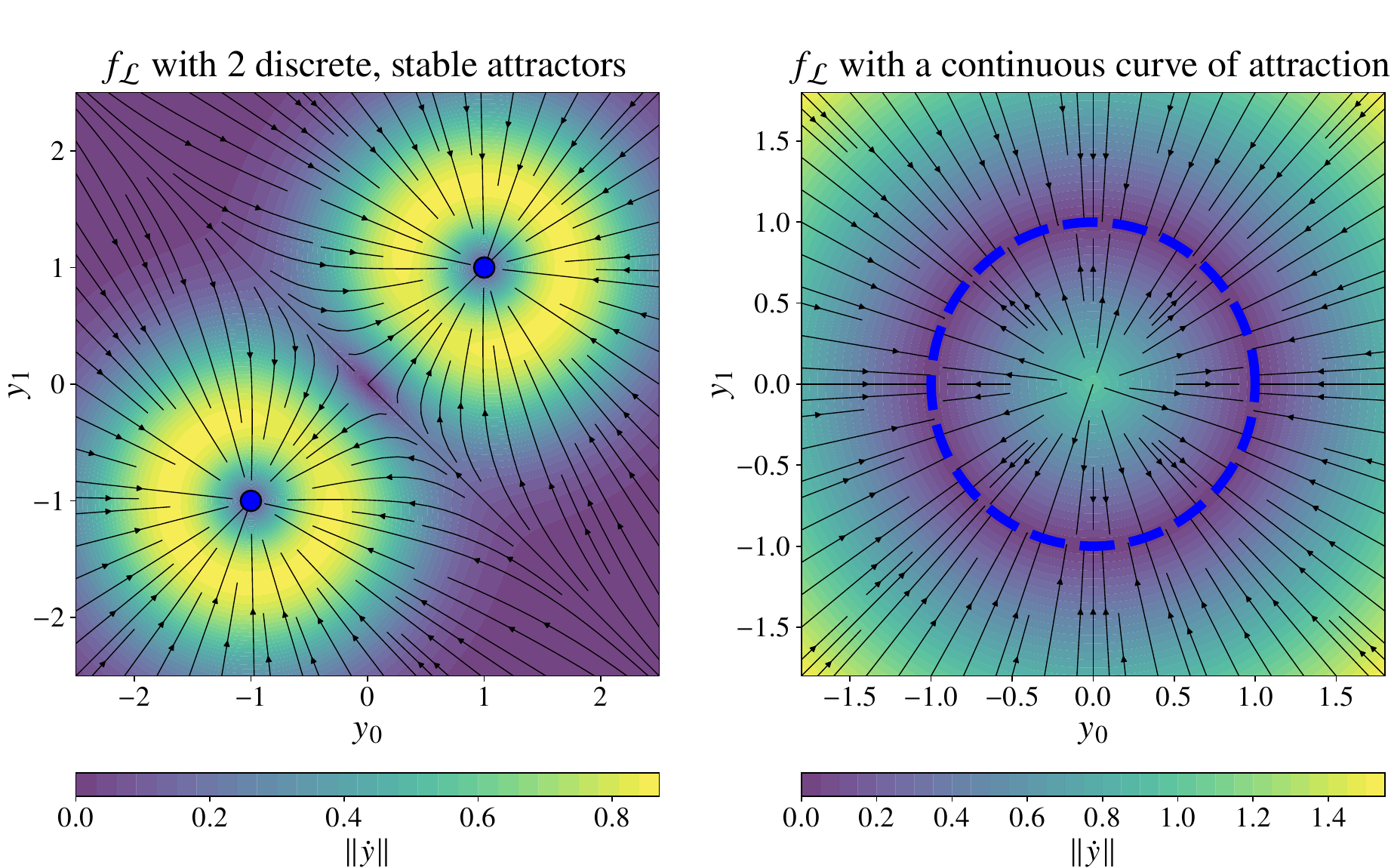}
    \caption{Reference dynamics $f_{\mathcal L}$ in two dimensions: two
    isolated stable attractors (blue dots, left) and a continuous attracting
    circle (dashed blue line, right).}
    \label{fig:potential_sys}
\end{figure}

Let $\mathcal{Z}=\{y:f_{\mathcal L}(y)=0\}$ denote the latent equilibrium set.
The following theorem transfers convergence to the task space through the
encoder; both reference systems introduced below admit a function
$V_{\mathcal L}$ satisfying its assumptions.
\begin{theorem}[Transfer of convergence]
\label{thm:transfer}
Let $\mathcal{T}$ be compact and positively invariant under $f_\theta$, and
let $V_{\mathcal L}$ be a $C^1$ function with
$\dot V_{\mathcal L}(y)=\nabla V_{\mathcal L}(y)^\top f_{\mathcal L}(y)\leq0$,
where $\dot V_{\mathcal L}(y)=0$ only if $f_{\mathcal L}(y)=0$. If the
dynamics-matching condition
\begin{equation}
J_{\psi_\theta}(x)\,f_\theta(x)=f_{\mathcal L}(\psi_\theta(x))
\quad\text{for all }x\in\mathcal{T},
\label{eq:dynamics_matching}
\end{equation}
holds, where $J_{\psi_\theta}=\partial\psi_\theta/\partial x$, then every
task trajectory converges to the set
$\mathcal{E}=\{x\in\mathcal{T}:\psi_\theta(x)\in\mathcal{Z}\}$.
\end{theorem}
\begin{proof}
Let $V_{\mathcal T}=V_{\mathcal L}\circ\psi_\theta$. By the chain rule and
\eqref{eq:dynamics_matching},
\begin{equation}
\begin{aligned}
\dot V_{\mathcal T}(x)
&=\nabla V_{\mathcal L}(\psi_\theta(x))^\top\,\frac{d\psi_\theta(x)}{dt}\\
&=\nabla V_{\mathcal L}(\psi_\theta(x))^\top\,\frac{\partial\psi_\theta(x)}{\partial x}\,\dot x\\
&=\nabla V_{\mathcal L}(\psi_\theta(x))^\top
\underbrace{J_{\psi_\theta}(x)\,f_\theta(x)}_{=\,f_{\mathcal L}(\psi_\theta(x))\text{ by \eqref{eq:dynamics_matching}}}\\
&=\dot V_{\mathcal L}(\psi_\theta(x))\leq0.
\end{aligned}
\label{eq:task_lyapunov_derivative}
\end{equation}
Since $\mathcal{T}$ is compact and positively invariant, LaSalle's invariance
principle~\cite{khalil2002NonlinearSystems} implies that every task trajectory
converges to the largest invariant set in $\{x:\dot V_{\mathcal T}(x)=0\}$.
Since $\dot V_{\mathcal T}(x)=\dot V_{\mathcal L}(\psi_\theta(x))=0$ if and
only if $\psi_\theta(x)\in\mathcal{Z}$, this set is contained in
$\mathcal{E}$.
\end{proof}
To connect this convergence result to the desired goals, we seek an encoding
that maps the goals onto the reference attractors. The reference dynamics
can also contain unstable equilibria, such as the saddle between the two
attractors in \autoref{fig:potential_sys}. Task states mapped onto these
equilibria also belong to $\mathcal{E}$. We therefore seek an encoding for
which the convergence set consists of the desired goals $\mathcal{G}$ and
task states whose latent representations coincide with unstable equilibria
of the reference dynamics,
\begin{equation}
\mathcal{U}=\bigl\{x\in\mathcal{T}\setminus\mathcal{G}:
\psi_\theta(x)\in\mathcal{Z}\setminus\psi_\theta(\mathcal{G})\bigr\}.
\label{eq:unstable_equilibrium_preimage}
\end{equation}
\begin{remark}[Practical effect of unstable equilibria]
Small perturbations along unstable directions drive trajectories away from
an unstable task-space equilibrium. In physical robot systems, measurement
noise and actuation or tracking errors typically introduce such
perturbations, making persistent trapping at a saddle point unlikely in
practice and allowing motion to continue toward a stable goal.
\end{remark}
This motivates three requirements on the encoder:
\begin{enumerate}
\item[(R1)] $\mathcal{G}\subseteq\mathcal{E}$: desired goals map to latent
equilibria;
\item[(R2)] $\mathcal{E}\subseteq\mathcal{G}\cup\mathcal{U}$: the convergence
set contains only the desired goals and task states mapped onto unstable
latent equilibria;
\item[(R3)] $\psi_\theta$ is one-to-one on $\mathcal{G} \cup \mathcal{U}$:
distinct goals and task states mapped onto unstable latent equilibria have
distinct latent representations.
\end{enumerate}
Under (R1) and (R2), $\mathcal{E}=\mathcal{G}\cup\mathcal{U}$, so every task
trajectory converges to this set. Requirement (R3) keeps goal encodings
distinct and limits $\mathcal{U}$ to at most one task state per unstable
latent equilibrium.
The matching loss \eqref{eq:match_loss} encourages the dynamics-matching
condition \eqref{eq:dynamics_matching}, while the separation loss
\eqref{eq:sep_loss} discourages collapsed encodings along trajectories (R2).
For discrete goals (Section~\ref{sec:discrete_goals}), the mapping loss
\eqref{eq:goal_mapping_loss} separates the Gaussian centers, promoting stable
equilibria near the goal encodings (R1) and keeping these encodings distinct
(R3). For goal contours (Section~\ref{sec:continuous_goals}), the curve loss
\eqref{eq:curve_loss} encourages goal encodings to lie on the attracting
circle (R1) and other states to lie away from it (R2), while the orientation
loss \eqref{eq:orient_loss} keeps contour goal encodings distinct (R3).
\subsection{Reference dynamics with discrete equilibria}
\label{sec:discrete_goals}
Let $y_g=\psi_\theta(x_g)$ be the latent encoding of task goal
$x_g\in\mathcal{G}$, $g=1,\dots,N_g$. We define the reference dynamics as
gradient ascent on a Gaussian potential:
\begin{equation}
P(y)=\sum_{g=1}^{N_g}e^{-\|y-y_g\|_2^2},
\quad
f_{\mathcal L}(y)=\nabla P(y).
\label{eq:multi_gaussian}
\end{equation}
With $V_{\mathcal L}=-P$, we obtain
$\dot V_{\mathcal L}=-\|f_{\mathcal L}\|_2^2\leq0$, which vanishes only at
latent equilibria, as Theorem~\ref{thm:transfer} requires. Trajectories are
bounded because the superlevel sets $\{y:P(y)\geq c\}$, $c>0$, are compact
and positively invariant.
In theory, closely spaced Gaussian centers can yield a single potential maximum. 
In practice, we prevent this by choosing fixed reference positions
$\xi_g$ with sufficient separation to support distinct attractors and
anchoring the goal encodings to these positions through the mapping loss,
\begin{equation}
\ell_{\mathrm{map}}
=
\frac{1}{N_g}\sum_{g=1}^{N_g}\|y_g-\xi_g\|_2^2.
\label{eq:goal_mapping_loss}
\end{equation}
This separation reduces the interaction between Gaussians and promotes
stable equilibria near the goal encodings, supporting (R1), while keeping
the goal encodings distinct, as required by (R3).
\subsection{Reference dynamics with a continuous set of attraction}
\label{sec:continuous_goals}
We represent the closed task-space goal contour
$\mathcal{G}=\mathcal{C}\subset\mathcal{T}$ by an attracting latent circle.
Partition the latent state as $y=(u,z)$, with $u\in\mathbb{R}^2$ and
$z\in\mathbb{R}^{N_y-2}$, and define the Cartesian reference dynamics
\begin{equation}
\dot u=\beta(1-\|u\|_2^2)u,
\qquad
\dot z=-z,
\label{eq:curve_field}
\end{equation}
where $\beta>0$. These dynamics are smooth everywhere, including at $u=0$.
For $r=\|u\|_2>0$, their polar interpretation is
$\dot r=\beta r(1-r^2)$ and $\dot\alpha=0$, where $\alpha$ is the latent
angle. Thus, for $u(0)\neq0$, the radius approaches one, the angle remains
constant, and $z$ decays to zero, yielding convergence to the latent goal set
\begin{equation}
S=\{(u,z):\|u\|_2=1,\ z=0\}.
\label{eq:latent_goal_circle}
\end{equation}
The function
\begin{equation}
\begin{aligned}
V_{\mathcal L}(y)&=\tfrac14(r^2-1)^2+\tfrac12\|z\|_2^2,\\
\dot V_{\mathcal L}(y)&=-\beta\,r^2(r^2-1)^2-\|z\|_2^2\leq0,
\end{aligned}
\label{eq:curve_lyapunov}
\end{equation}
satisfies the assumptions of Theorem~\ref{thm:transfer}: its derivative
vanishes only on the equilibrium set $S\cup\{0\}$. Its compact, positively
invariant sublevel sets ensure bounded trajectories. Initial states with
$u(0)=0$ converge to the unstable origin, whose task-space preimage forms
$\mathcal{U}$.
To place contour encodings at unit radius and separate non-goal encodings
from the latent goal set $S$, we introduce a curve loss. We sample contour points
$\mathcal{B}_{\mathcal C}\subset\mathcal{C}$ uniformly along the curve and
form $\mathcal{B}_{\mathrm{neg}}$ from non-goal points sampled both near the
contour on either side and uniformly throughout the workspace. Write
$r(x)=\|u(\psi_\theta(x))\|_2$ and $z(x)=z(\psi_\theta(x))$. The squared
Euclidean distance from $\psi_\theta(x)$ to $S$ is
$d_S(x)^2=(r(x)-1)^2+\|z(x)\|_2^2$. We define
\begin{equation}
\begin{aligned}
\ell_{\mathrm{curve}}
={}&\frac{1}{|\mathcal{B}_{\mathcal C}|}\sum_{x\in\mathcal{B}_{\mathcal C}}
\bigl(r(x)-1\bigr)^2\\
&+\frac{1}{|\mathcal{B}_{\mathrm{neg}}|}\sum_{x\in\mathcal{B}_{\mathrm{neg}}}
\max\bigl(0,\,\epsilon_c-d_S(x)^2\bigr).
\end{aligned}
\label{eq:curve_loss}
\end{equation}
The first term encourages unit radius for contour encodings, while the
matching loss \eqref{eq:match_loss} encourages their remaining coordinates
to follow $\dot z=-z$ and vanish as the task motion settles at a goal.
Together, these losses encourage goal encodings to lie on $S$, supporting
(R1). The second term penalizes sampled non-goal encodings within a distance
$\sqrt{\epsilon_c}$ of $S$, supporting (R2) alongside the separation loss
\eqref{eq:sep_loss}.
Radial placement leaves the angular arrangement of contour encodings
unspecified. To distinguish contour points, we order them along the curve
and assign each $x_i$ a target angle $\hat\alpha_i\in[-\pi,\pi)$ during
preprocessing. With target unit vectors
$\hat u_i=(\cos\hat\alpha_i,\sin\hat\alpha_i)^\top$, the orientation loss is
\begin{equation}
\ell_{\mathrm{orient}}
=\frac{1}{|\mathcal{B}_{\mathcal C}|}\sum_{x_i\in\mathcal{B}_{\mathcal C}}
\left\|\frac{u(\psi_\theta(x_i))}{\max(r(x_i),\varepsilon)}-\hat u_i\right\|_2^2,
\label{eq:orient_loss}
\end{equation}
where $\varepsilon>0$ is a small normalization constant that keeps the
expression defined at $r(x_i)=0$. Comparing unit-vector targets accounts for
angle periodicity and encourages distinct contour goal encodings,
supporting (R3).
\subsection{Imitation loss and training objective}
\label{sec:imitation_loss}
Following CONDOR~\cite{perezDattari2023Condor}, we encourage reproduction of the
demonstrations through a multistep imitation loss. We sample $B^i$
demonstration segments with states $x^*_{t,b}$ and roll out the task dynamics
for $H^i$ steps from $\hat x_{0,b}=x^*_{0,b}$:
\begin{equation}
\hat x_{t,b}=\hat x_{t-1,b}+\Delta t\,f_\theta(\hat x_{t-1,b}).
\label{eq:imitation_rollout}
\end{equation}
The imitation loss is the mean squared trajectory error:
\begin{equation}
\ell_{\mathrm{IL}}=\frac{1}{B^iH^i}
\sum_{b=0}^{B^i-1}\sum_{t=1}^{H^i}\bigl\|x^*_{t,b}-\hat x_{t,b}\bigr\|_2^2.
\label{eq:il_loss}
\end{equation}
We combine imitation, stability, and reference-specific losses into the
training objectives for discrete goals,
\begin{equation}
\mathcal{L}_{\mathrm{d}}
=
\lambda_{\mathrm{IL}}\ell_{\mathrm{IL}}
+\lambda_{\mathrm{stable}}\ell_{\mathrm{stable}}
+\lambda_{\mathrm{map}}(k)\,\ell_{\mathrm{map}},
\label{eq:discrete_training_objective}
\end{equation}
and continuous goals,
\begin{equation}
\begin{aligned}
\mathcal{L}_{\mathrm{c}}
={}&\lambda_{\mathrm{IL}}\ell_{\mathrm{IL}}
+\lambda_{\mathrm{stable}}\ell_{\mathrm{stable}} +\lambda_{\mathrm{curve}}\ell_{\mathrm{curve}}
+\lambda_{\mathrm{orient}}\ell_{\mathrm{orient}}.
\end{aligned}
\label{eq:continuous_training_objective}
\end{equation}
We observe empirically that goal separation is maintained after the initial
training iterations, so we progressively reduce the mapping-loss weight as
\begin{equation}
\lambda_{\mathrm{map}}(k+1)=\gamma\lambda_{\mathrm{map}}(k),
\label{eq:mapping_weight_schedule}
\end{equation}
where $k$ indexes training iterations and $\gamma\in(0,1)$.
We jointly optimize the network parameters $\theta$, with nonnegative
weights balancing the loss terms.
\section{Results}
We first benchmark multi-CONDOR on tasks with discrete goals, then
demonstrate its application to robotic grasping and illustrate its
extension to continuous goal contours.
\subsection{Discrete multi-attractor benchmarking}
\label{subsec:benchmarks}
\subsubsection{Datasets}
We evaluate multi-CONDOR on seven 2D motion tasks drawn manually on a
digital interface: three with two attractors and four with three attractors.
The tasks contain 6--8 and 12--24 demonstrations, respectively, with equal
numbers of demonstrations per goal. Goal positions are estimated by
clustering demonstration endpoints using $k$-means, with $k$ equal to the
number of attractors, and taking each cluster's mean endpoint.
All states are normalized to a workspace defined by the demonstration
minima and maxima multiplied by a factor of $1.3$ to provide a margin.
\subsubsection{Metrics}
We evaluate imitation accuracy and empirical stability. For imitation
accuracy, we simulate each system from the demonstration initial states
and compare the resulting trajectories with the demonstrations using root
mean squared error (RMSE), dynamic time warping distance normalized by the
number of points (DTWD) \cite{sakoe1978dynamic}, and Fr\'echet distance (FD) \cite{eiter1994computing}.
To assess stability, we simulate each system for $5000$ steps from a
$16\times16$ grid covering the workspace. We report the percentage of
initial states whose final state lies more than $5$ units (approximately
1\% of the workspace width) from every goal as the ``spurious attractors''
metric. We also report the mean distance from each final state to its
nearest goal (distance from goals).
\subsubsection{Baselines}
We compare our method with two baselines.
\emph{Single-attractor CONDORs + state machine} trains one
CONDOR model~\cite{perezDattari2023Condor} per goal and selects the nearest goal's model at
execution time, mirroring the state-space partitioning idea of
Augmented-SVM~\cite{shukla2012AugmentedSvm}. Each model combines imitation learning
with a contrastive stability loss that encourages encoded trajectories to
follow reference dynamics with a single stable equilibrium.
Each model uses the same architecture and optimizer as multi-CONDOR,
specified below.
\emph{StableNODE}~\cite{sochopoulos2024StableNeuralOdes}, our main baseline for learning a single
autonomous dynamical system with multiple goals, uses a neural ODE with a
Lyapunov-based correction to its nominal vector field. The correction acts when the
nominal dynamics violate a prescribed Lyapunov decrease condition. Its
formulation admits nonconvex Lyapunov functions, allowing multiple
attractors. We train it using the average Hausdorff distance between
generated and demonstrated trajectories, which measures geometric
similarity without requiring temporal alignment. The reported results use
$\alpha=\epsilon=10^{-2}$, where $\alpha$ specifies the desired Lyapunov
decay rate and $\epsilon$ regularizes the correction near points where the
Lyapunov gradient vanishes.
\subsubsection{Ablations}
We evaluate two ablations of the discrete-goal objective
\eqref{eq:discrete_training_objective}, retaining the multi-CONDOR network
architecture.
\emph{Only imitation} removes the stability loss (trajectory matching and
separation) and the goal-mapping loss
($\lambda_{\mathrm{stable}}=\lambda_{\mathrm{map}}=0$). It tests whether
fitting demonstrations alone yields accurate imitation and reliable goal
reaching throughout the workspace.
\emph{No mapping loss} removes only the goal-mapping loss
($\lambda_{\mathrm{map}}=0$), retaining imitation and stability losses.
It tests whether anchoring goal encodings to separated reference positions
$\xi_g$ improves imitation accuracy and suppresses spurious attractors.
\subsubsection{Multi-CONDOR implementation details}
\label{sec:implementation}
As a safeguard, rollout transitions that would leave $\mathcal{T}$ are
projected back onto its boundary.
The encoder has four hidden layers and the decoder three, all with $300$
units, GELU activations and layer normalization, and the latent space has
dimension $300$. 
The reference system is the Gaussian mixture of
Section~\ref{sec:discrete_goals}. 
For two attractors, the reference positions are
$\xi_1=(1,\dots,1)$ and $\xi_2=(-1,\dots,-1)$; for three attractors, they are
sampled uniformly at random in the latent space.
We use horizons $H^i=10$ and $H^s=7$, margin $m=0.033$, weights $\lambda_{\mathrm{IL}}=1$, $\lambda_{\mathrm{stable}}=0.4$ and
$\lambda_{\mathrm{map}}=0.05$ with decay factor $\gamma=0.99$, and train for
$10{,}000$ iterations with AdamW (learning rate $4.9\times10^{-4}$, weight
decay $10^{-5}$, batch size $250$).
All results are averaged over five random seeds.
\subsubsection{Discussion}
\autoref{tab:partitioned_performance_metrics} summarizes the results.
The imitation-only ablation achieves the lowest imitation errors in both
settings, but its spurious-attractor rates are approximately $17\%$ and
$37\%$ for two and three attractors, respectively. The full method records
no spurious attractors under our evaluation criterion, with modestly higher
imitation errors. Its mean distance from the goals decreases from $13.33$
to $3.10$ and from $28.24$ to $2.69$, respectively. Removing the mapping loss
also worsens both stability metrics in both settings and substantially
increases imitation errors for two attractors.
%
\input{comparison_tab}

\begin{figure}
    \centering
    \includegraphics[width=0.95\linewidth]{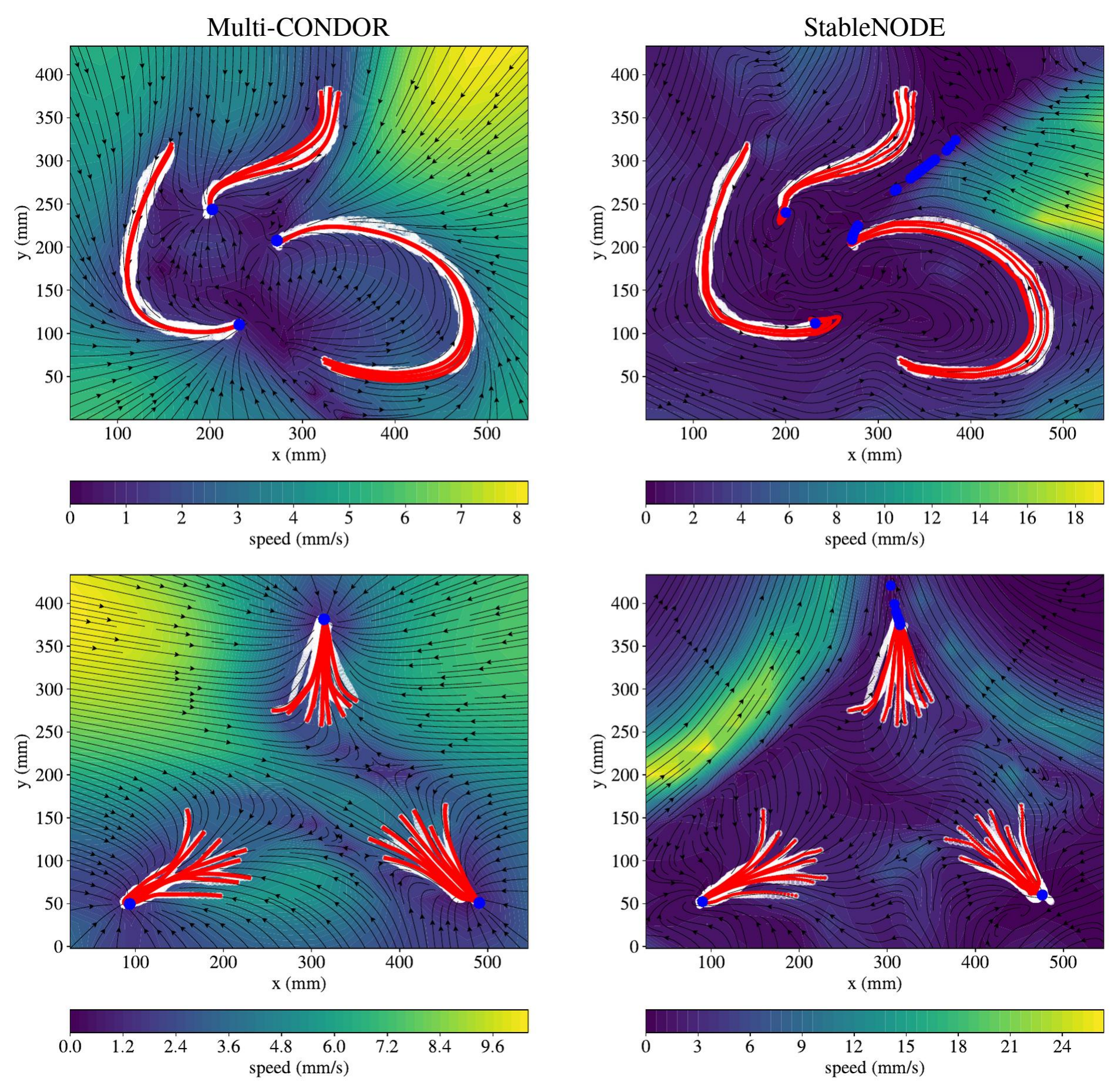}
    \caption{Learned dynamics of multi-CONDOR (left) and
    StableNODE~\cite{sochopoulos2024StableNeuralOdes} (right) on two three-attractor benchmark
    tasks. Demonstrations are white, generated trajectories red, and
    attractors blue. StableNODE produces additional spurious attractors
    away from the intended goals.}
  \label{fig:comparison_stablenode}
\end{figure}

Compared with separate single-attractor CONDOR models selected by a state
machine, multi-CONDOR achieves substantially lower imitation errors and
better stability metrics in both settings. StableNODE achieves comparable
imitation accuracy and the smallest mean distance from the goals for two
attractors. For three attractors, however, its spurious-attractor rate is
$7.29\%$, and its mean distance from the goals is $17.90$, compared with
$2.69$ for multi-CONDOR.
During StableNODE hyperparameter tuning and longer training runs, we
observed a tradeoff between imitation accuracy and stability: improving
one often degraded the other.
\autoref{fig:comparison_stablenode} illustrates this qualitatively:
StableNODE produces additional spurious attractors (blue) away from the
demonstration endpoints, with demonstrations shown in white and generated
trajectories in red. Overall, multi-CONDOR combines imitation accuracy
close to the imitation-only ablation with reliable goal reaching across
the evaluated initial states.
\subsection{A multi-goal robotic grasping task}
\begin{figure}[hbtp!]
    \centering
    \includegraphics[width=0.23\textwidth]{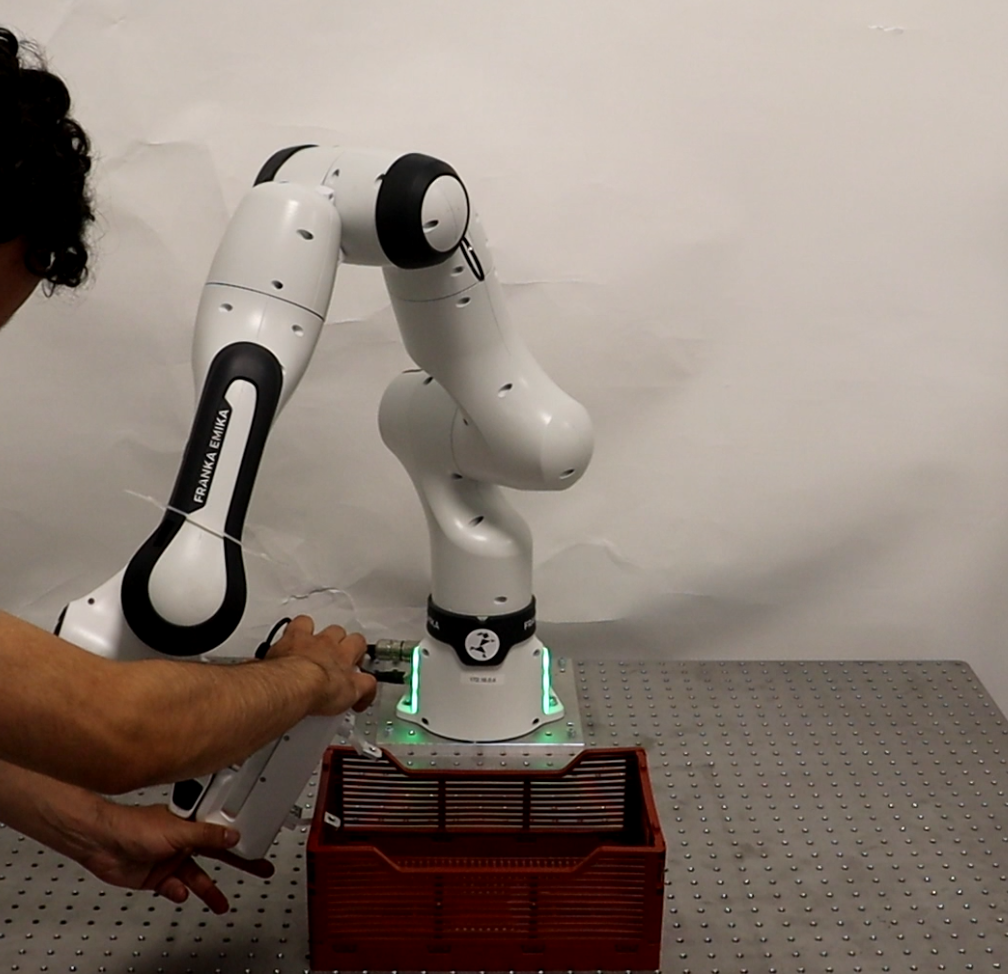}
    \includegraphics[width=0.23\textwidth]{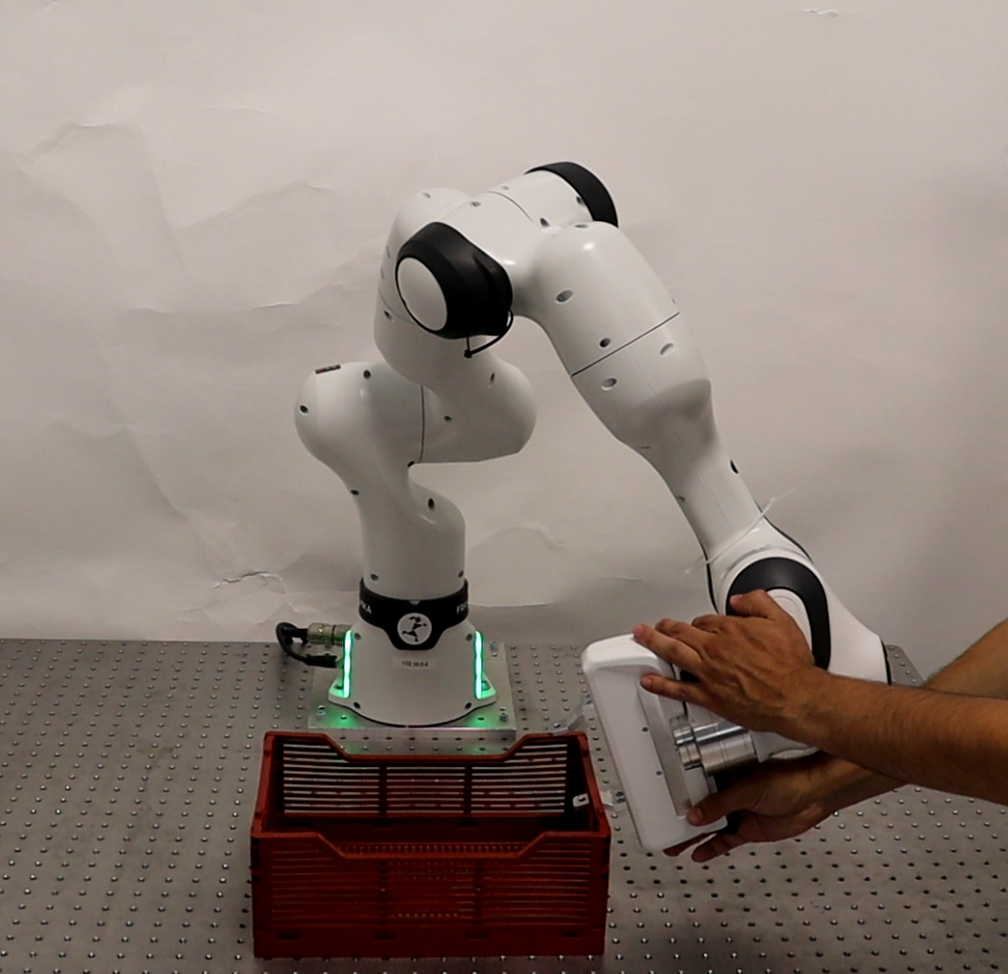}
    \caption{Kinesthetic demonstrations approaching the two grasping goals.}
    \label{fig:robot_demonstrations}
\end{figure}

\begin{figure}[hbtp!]
    \centering
    \includegraphics[width=0.23\textwidth]{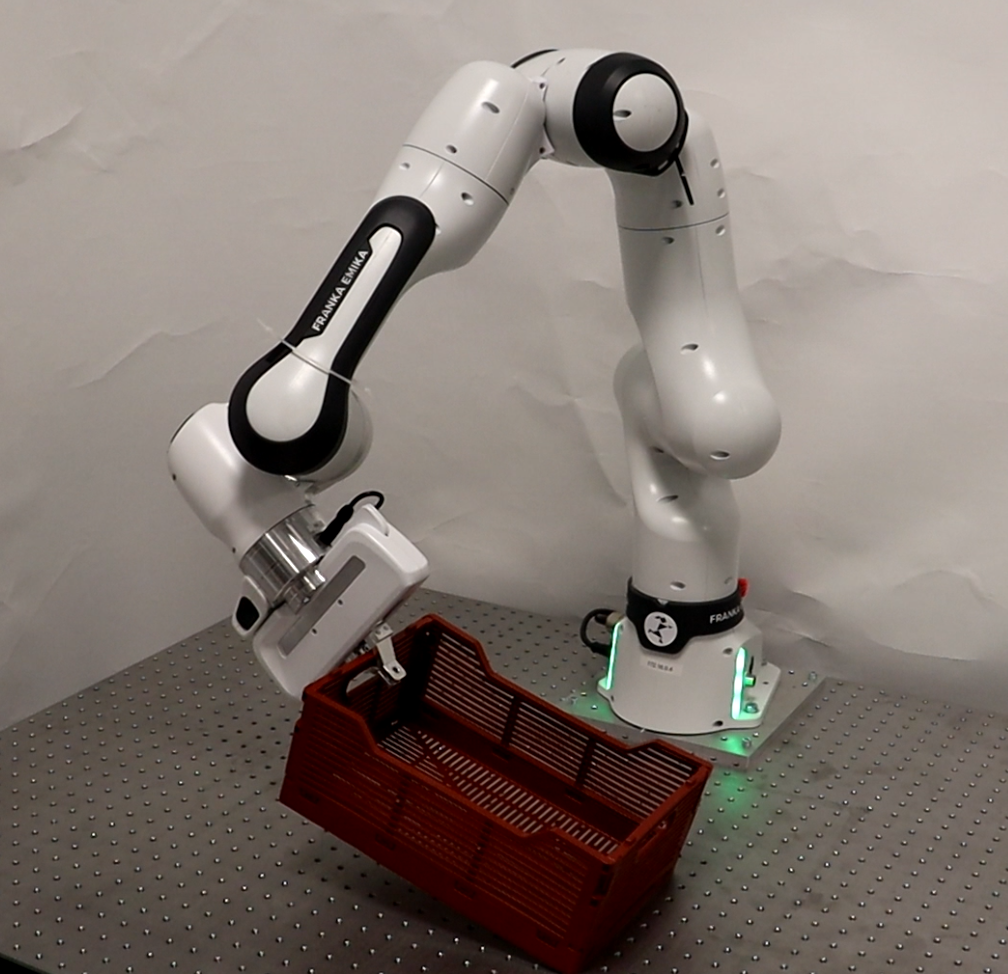}
    \includegraphics[width=0.23\textwidth]{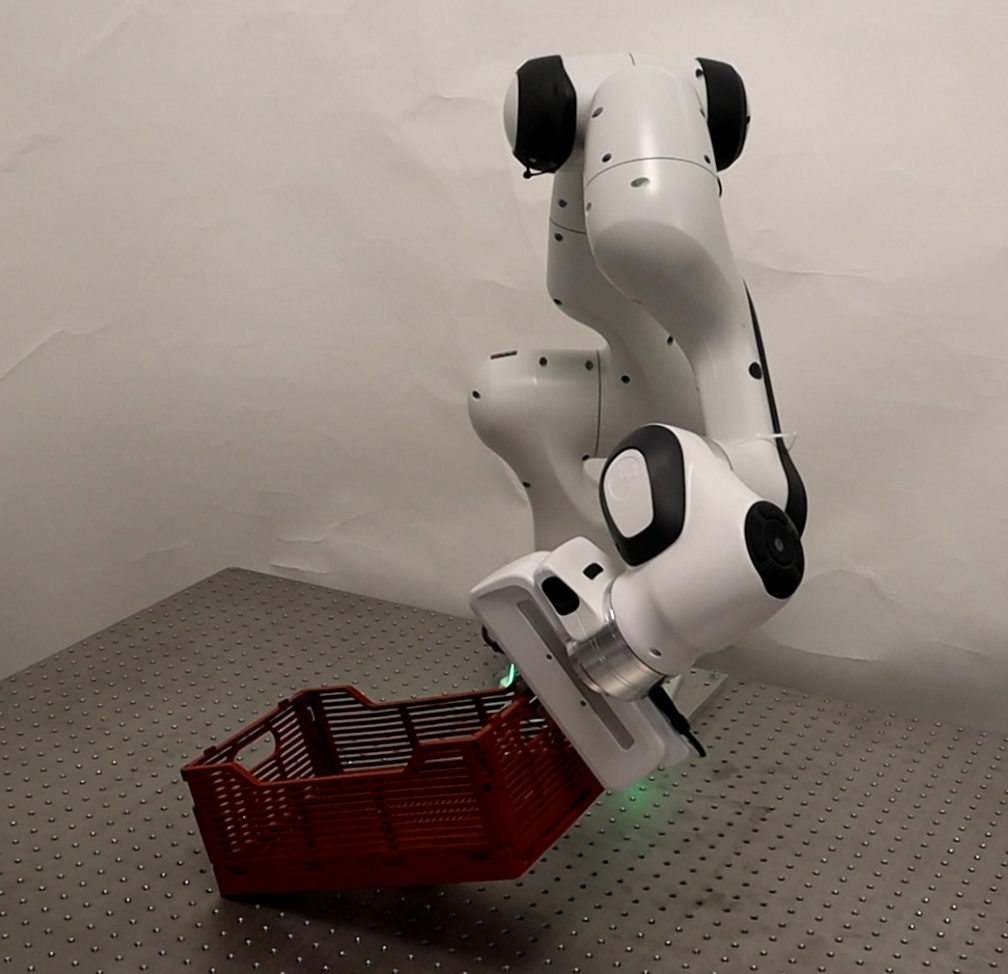}
    \caption{Robot executions reaching either grasping goal using a single
    vector field learned with multi-CONDOR.}
    \label{fig:robot_executions}
\end{figure}

\begin{figure}
    \centering
    \includegraphics[width=0.8\linewidth, trim=0cm 1cm 0cm 1cm, clip]{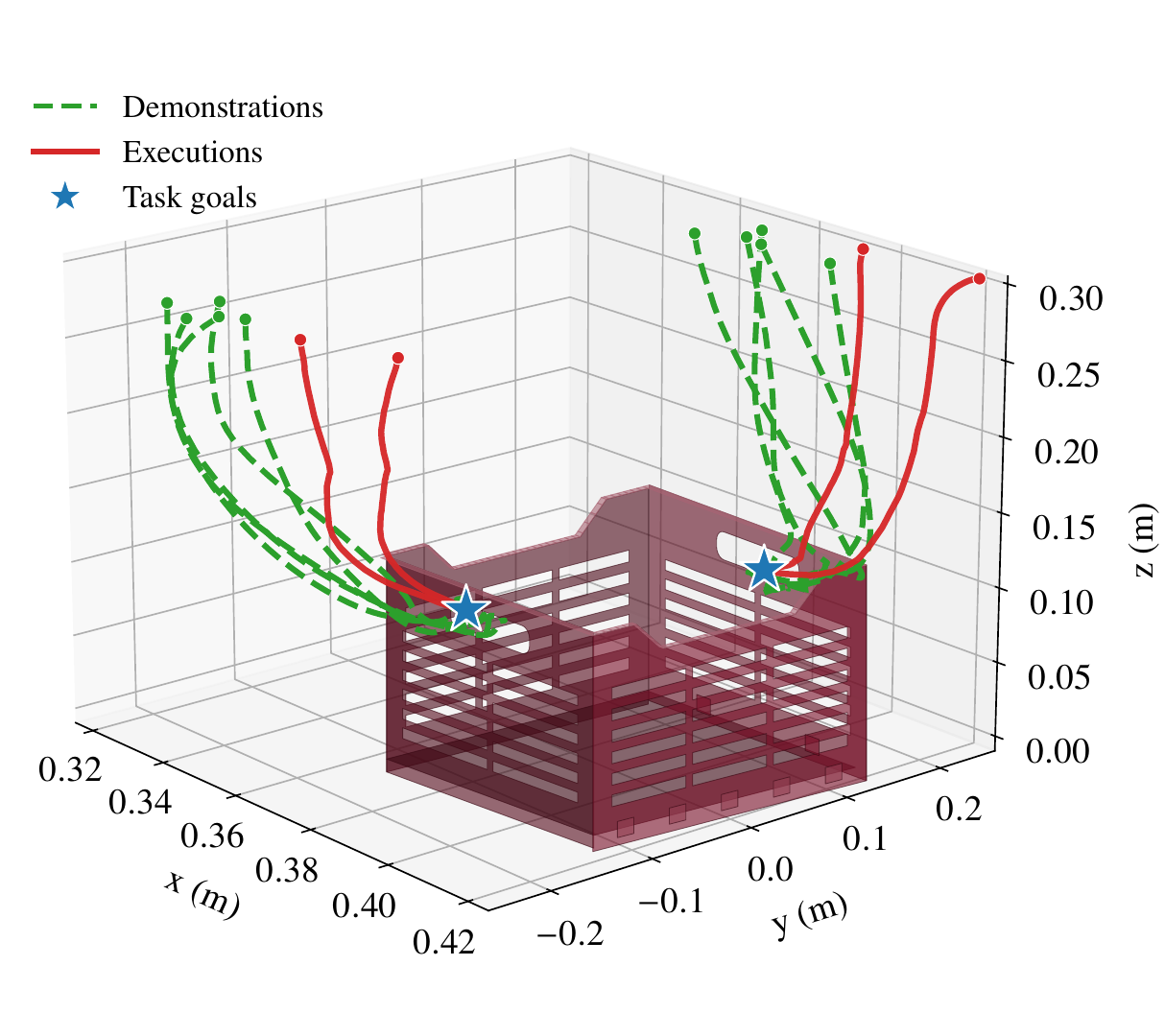}
    \caption{Visualization of provided demonstrations (in green) for the multi-goal grasping task and four multi-CONDOR trajectories (in red) which successfully grasped the box by the handles. Note how the trajectories converge to the desired goals despite starting from initial conditions which are not the same as the provided demonstrations.}
    \label{fig:demos_execs_crate}
\end{figure}

We evaluate multi-CONDOR on a robotic grasping task in which a Franka FR3 grasps a box from either side. 
We collect ten kinesthetic demonstrations, five per goal, and learn dynamics for the 3D Cartesian end-effector
position (\autoref{fig:robot_demonstrations}).
At each control step, the learned dynamics map the current position to a velocity, which is integrated to obtain the target position for an impedance controller. 
The target orientation is taken from the nearest recorded demonstration point, and a separate state machine closes the gripper when the end effector reaches either goal.
\autoref{fig:robot_executions} shows the robot reaching either goal depending on its initial position.
\autoref{fig:demos_execs_crate} shows the 5 provided demonstrations for each side in green, and 4 examples of successful trajectories in red.
Note how the trajectories converge towards the desired goal despite the fact that they do not start at the same initial conditions as the provided demonstrations. 
Moreover, to be successful, these trajectories had to follow the demonstrations in approaching from the side, rather than from the top, to be able to insert one of the fingers in the handle.
This task thus highlights how the system can converge towards the goal when starting from varying initial conditions, while still imitating the demonstrated behaviour when necessary. 
\vspace{-3mm}
\subsection{Convergence to closed curves}
\begin{figure}
    \centering
    \includegraphics[width=0.47\linewidth]{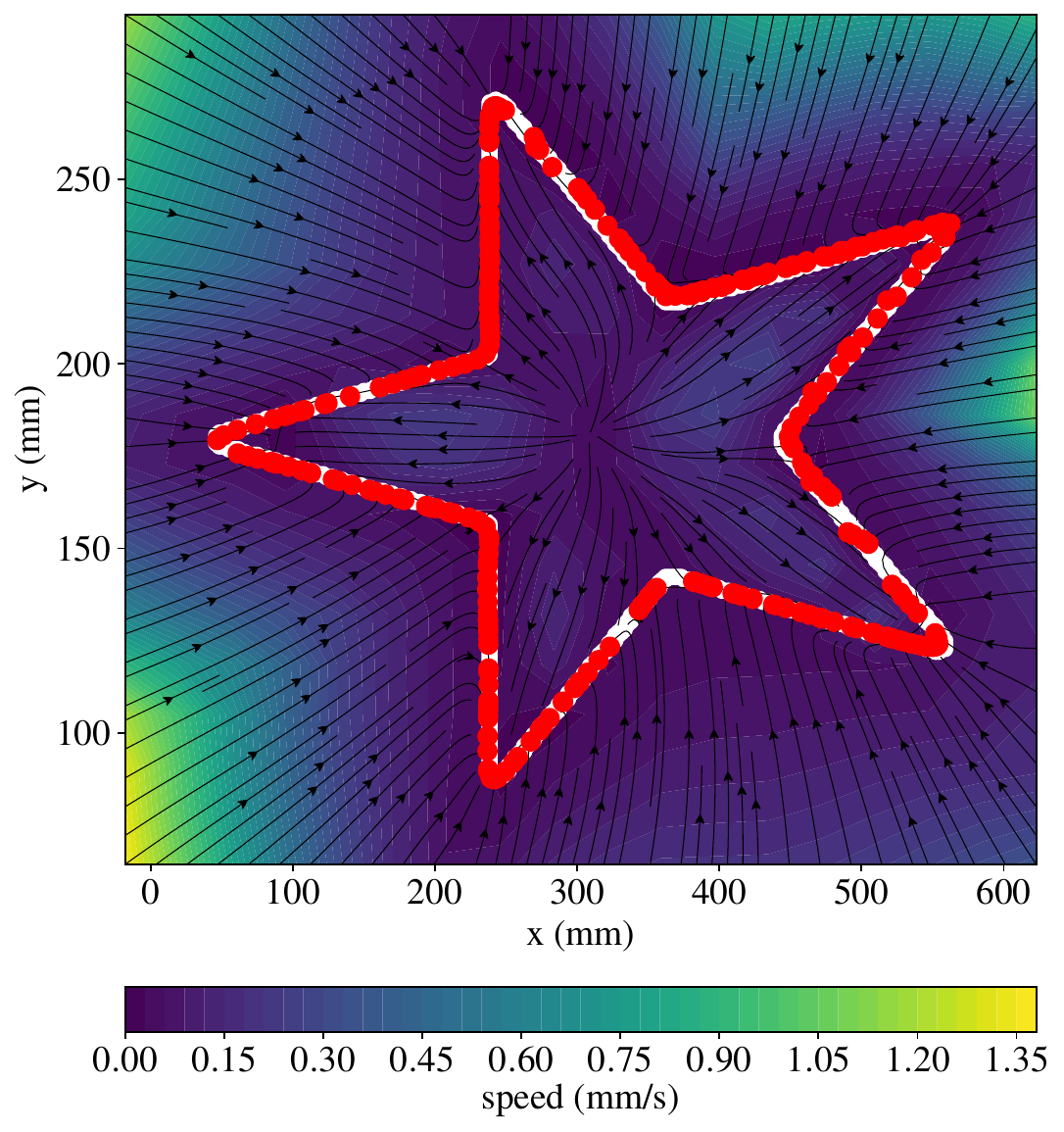}
    \includegraphics[width=0.47\linewidth]{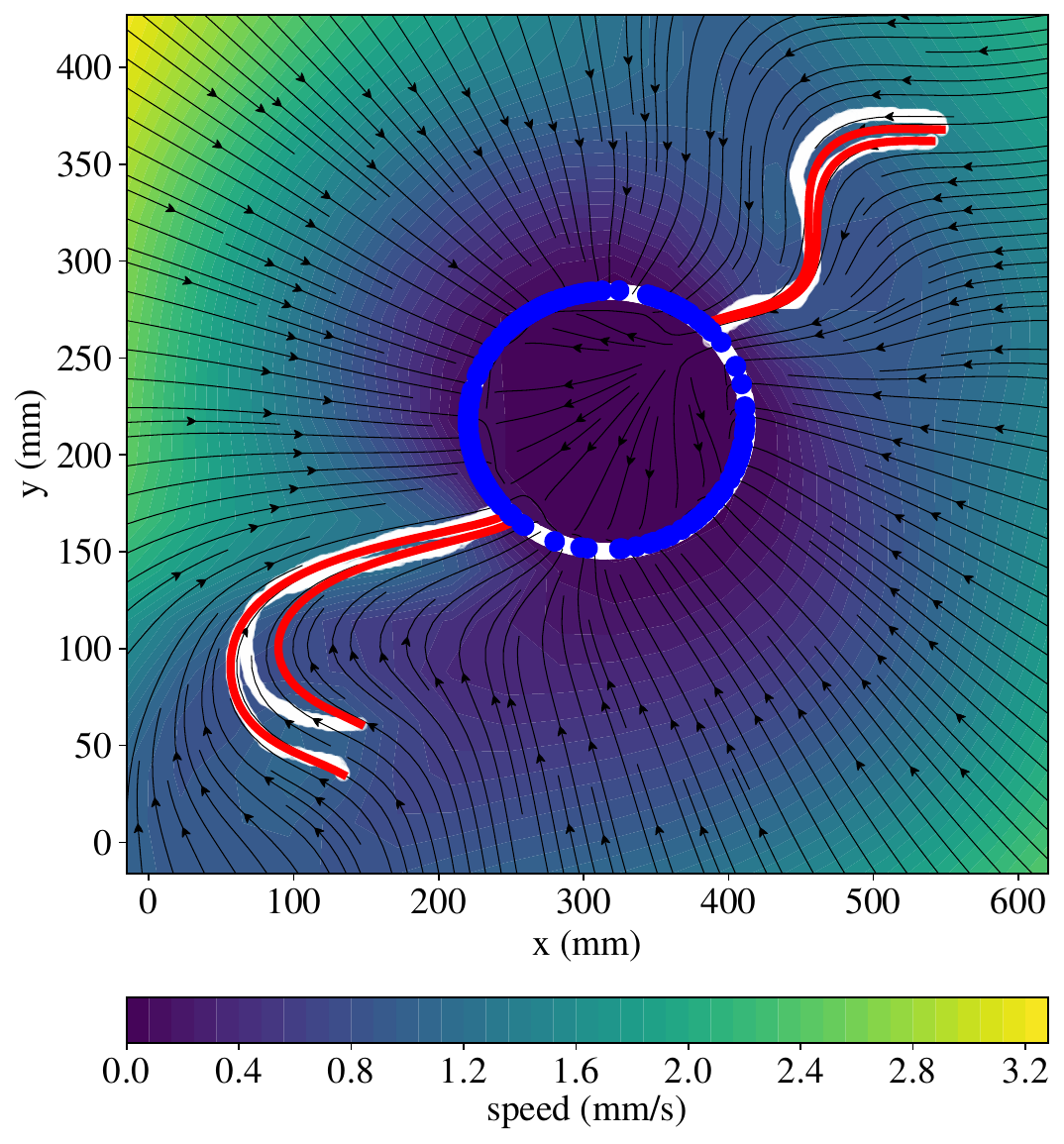}
    \caption{Example of learned dynamics using multi-CONDOR for a continuous, closed star shape (left) without any imitation, and an example of a case with a circular shape as a continuous goal with two subgroups of demonstrations to imitate (right).}
    \label{fig:continous_star}
\end{figure}

We also provide qualitative results which use the reference dynamics with a continuous curve of attraction. 
In this case, rather than task dynamics which reach multiple discrete goals, we can learn dynamics which converge to a continuous, closed shape.
As \autoref{fig:continous_star} shows, for example, we can learn to converge to a star-shaped contour: the red attractors shown on the image on the left all lie on the star-shaped goal contour.
Moreover, we can still combine imitation with this approach, as the plot on the right shows with the demonstrations in white, trajectories in red, and the attractors lying on the target circle in blue.
Similarly to the previous grasping task, such dynamics could thus be useful when an object can be grasped along an entire continuous set of points, rather than specific, distinct ones. 
\section{Conclusion, limitations, and future work}
Grasping a container by either handle or at admissible points along its rim exemplifies manipulation tasks with multiple valid endpoints.
We presented a CONDOR-based framework for learning such motions toward discrete goals or continuous attracting goal contours.
Our method reduces mean trajectory RMSE by approximately $72\%$ relative to single-attractor CONDORs with a state machine.
The observed rate of spurious attractors falls from StableNODE's $0.31\%$ and $7.29\%$ to $0\%$ for two and three goals, respectively.
A robotic grasping demonstration further illustrates reaching alternative goals within a single autonomous vector field.

Our Lyapunov analysis establishes sufficient conditions for transferring convergence from the reference dynamics to the learned task dynamics.
These guarantees require exact dynamics matching, which training only encourages.
A diffeomorphic encoder, following Euclideanizing Flows~\cite{rana2020EuclideanizingFlows}, with task velocities derived through its inverse Jacobian could enforce this condition by construction.
The analysis also requires the intended correspondence between goals and latent attractors, including a distinct stable equilibrium for each discrete goal.
Since Gaussian interactions can shift or merge equilibria, this extension should incorporate latent dynamics with distinct stable equilibria fixed at the encoded discrete goals.

The current formulation treats discrete attractors and continuous attracting contours separately.
Future work could unify these with single point attractors and limit cycles, with convergence guarantees appropriate to each attractor type.
\vspace{-1mm}
\section*{Acknowledgements}
The illustrations of the human demonstrator in \autoref{fig:method_summary} were generated using OpenAI's ChatGPT image generation. Claude Opus~5.5 (Anthropic) assisted in writing code for plots and TikZ drawings, Claude Opus~5.5 and GPT-6 Astra were also used to edit and revise the manuscript. The authors reviewed and edited all AI-assisted content and take full responsibility for the content of this paper.
\bibliographystyle{IEEEtran}
\bibliography{ref}

\end{document}

%% file: tikz_preamble.tex
\usepackage{tikz}
\usetikzlibrary{arrows.meta,decorations.markings,calc}

\definecolor{taskLine}{HTML}{2E75B6}  \definecolor{taskDot}{HTML}{1F3864}
\definecolor{latLine}{HTML}{8CC56B}   \definecolor{latDot}{HTML}{00A651}
\definecolor{mapLine}{HTML}{E2582F}   \definecolor{mapDot}{HTML}{D90000}
\definecolor{startCol}{HTML}{FFC000}
\definecolor{boxTask}{HTML}{CFE0F1}
\definecolor{boxLat}{HTML}{D5E8C8}
\definecolor{boxMap}{HTML}{F8B4B4}

\tikzset{
  pt/.style={circle, draw=black!55, fill=#1, line width=0.5pt, inner sep=0pt, minimum size=7.5pt},
  traj/.style={line width=1.6pt, line cap=round},
  b0/.style={},                                   
  b1/.style={dash pattern=on 7pt off 3.5pt},      
  tail/.style={line width=1.6pt, line cap=round, dash pattern=on 0pt off 3.6pt},
  match/.style={{Stealth[length=5.5pt,width=5pt]}-{Stealth[length=5.5pt,width=5pt]},
               line width=0.9pt, black!75, shorten <=5pt, shorten >=5pt},
  lbl/.style={inner sep=1pt, font=\large},
  sysbox/.style={inner xsep=6pt, inner ysep=4pt, font=\Large},
}

\newcommand{\Traj}[4]{%
  \draw[traj,#2,postaction={decorate,decoration={markings,
      mark=at position 0     with {\coordinate (#1-0);},
      mark=at position 0.3333 with {\coordinate (#1-1);},
      mark=at position 0.6667 with {\coordinate (#1-2);},
      mark=at position 0.999  with {\coordinate (#1-3);}}}] #3;
  \draw[tail,#2,tail,-{Triangle[length=8pt,width=8pt]},shorten >=2.5pt] (#1-3) #4 coordinate (#1-N);
}

%% file: matching_tikz.tex
\begin{tikzpicture}[x=0.87cm,y=0.87cm] 
\path (0,0) rectangle (17.2,10.2);
\draw[line width=1.8pt, dash pattern=on 5pt off 3pt]
  (0.1,3.3) to[out=-5,in=195]  (6.4,4.6)
            to[out=15,in=180]  (7.2,4.85)
            to[out=0,in=170]   (8.0,4.7)
            to[out=-10,in=215] (10.6,5.45)
            to[out=35,in=240]  (12.9,8.4)
            to[out=60,in=235]  (13.8,10.0);

\node[font=\LARGE] at (2.4,9.4)  {Task Space};
\node[font=\LARGE] at (8.1,0.6) {Latent Space};

\Traj{x0}{taskLine,b0}
  {(0.9,7.5) to[out=-5,in=125] (2.5,6.8) to[out=-55,in=155] (3.9,5.75)}
  {to[out=-25,in=165] (4.9,5.4)}
\Traj{x1}{taskLine,b1}
  {(10.7,7.0) to[out=100,in=0] (9.3,8.5) to[out=180,in=70] (7.9,7.3)}
  {to[out=-110,in=85] (7.7,6.3)}

\coordinate (s0) at (4.9,2.4);
\coordinate (g1) at (9.2,1.7);
\Traj{l0}{latLine,b0}
  {(s0) -- ($(s0)!0.74!(g1)$)}
  {-- (g1)}
\Traj{m0}{mapLine,b0}
  {(s0) to[out=45,in=210] (6.2,3.2) to[out=25,in=200] (8.2,3.75)}
  {to[out=20,in=195] (9.3,4.05)}

\coordinate (s1) at (14.2,5.6);
\coordinate (g2) at (13.1,1.8);
\Traj{l1}{latLine,b1}
  {(s1) -- ($(s1)!0.74!(g2)$)}
  {-- (g2)}
\Traj{m1}{mapLine,b1}
  {(s1) to[out=-10,in=95] (15.8,4.0) to[out=-85,in=45] (15.3,2.45)}
  {to[out=-135,in=65] (14.7,1.6)}

\draw[match] (l0-2) -- (m0-2) node[midway,right=1pt,font=\large] {$\ell_{\mathrm{match}}$};
\draw[match] (l1-2) -- (m1-2) node[midway,above=1pt,font=\large] {$\ell_{\mathrm{match}}$};

\foreach \i/\a in {0/above,1/above right,2/above right,3/above right}
  \node[pt=taskDot] at (x0-\i) {} node[lbl,\a=2pt] at (x0-\i) {$x_{\i,0}$};
\node[pt=taskDot] at (x0-N) {} node[lbl,above right=1pt] at (x0-N) {$x_{H^s\!,0}$};

\foreach \i/\a in {0/right,1/above right,2/above left,3/left}
  \node[pt=taskDot] at (x1-\i) {} node[lbl,\a=2pt] at (x1-\i) {$x_{\i,1}$};
\node[pt=taskDot] at (x1-N) {} node[lbl,left=3pt] at (x1-N) {$x_{H^s\!,1}$};

\foreach \i in {1,...,3}
  \node[pt=latDot] at (l0-\i) {} node[lbl,below=3pt] at (l0-\i) {$y^{\mathcal L}_{\i,0}$};
\node[pt=latDot] at (l0-N) {} node[lbl,above right=1pt] at (l0-N) {$y_{g_1}=\psi_\theta(x_{g_1})$};
\foreach \i in {1,...,3}
  \node[pt=mapDot] at (m0-\i) {} node[lbl,above left=1pt] at (m0-\i) {$y^{\mathcal T}_{\i,0}$};
\node[pt=mapDot] at (m0-N) {} node[lbl,right=3pt] at (m0-N) {$y^{\mathcal T}_{H^s\!,0}$};
\node[pt=startCol] at (s0) {};
\node[lbl,left=4pt] at (s0)
  {$y^{\mathcal T}_{0,0}=y^{\mathcal L}_{0,0}=\psi_\theta(x_{0,0})$};

\foreach \i in {1,...,3}
  \node[pt=latDot] at (l1-\i) {} node[lbl,left=3pt] at (l1-\i) {$y^{\mathcal L}_{\i,1}$};
\node[pt=latDot] at (l1-N) {} node[lbl,below=4pt] at (l1-N) {$y_{g_2}=\psi_\theta(x_{g_2})$};
\foreach \i/\a in {1/above right,2/right,3/right}
  \node[pt=mapDot] at (m1-\i) {} node[lbl,\a=5pt] at (m1-\i) {$y^{\mathcal T}_{\i,1}$};
\node[pt=mapDot] at (m1-N) {} node[lbl,right=5pt] at (m1-N) {$y^{\mathcal T}_{H^s\!,1}$};
\node[pt=startCol] at (s1) {};
\node[lbl,anchor=south,xshift=6pt,yshift=5pt] at (s1)
  {$y^{\mathcal T}_{0,1}=y^{\mathcal L}_{0,1}=\psi_\theta(x_{0,1})$};

\node[sysbox, fill=boxTask] at (5.8,7.6) {$f^{\mathcal T}_{\theta}(x)$};
\node[sysbox, fill=boxLat]  at (1.4,1.1) {$f_{\mathcal L}(y)$};
\node[sysbox, fill=boxMap]  at (4.1,1.1) {$\psi_\theta(x_{t,b})$};

\end{tikzpicture}%
 

%% file: comparison_tab.tex
\begin{table*}[htbp!]
\vspace{5pt}
\centering
\caption{Performance metrics for the 2-attractor and 3-attractor datasets. We report the mean and standard deviation over the total number of tasks (3 for 2-attractor and 4 for 3-attractor) and over 5 random seeds. The best results are shown in bold.}
\label{tab:partitioned_performance_metrics}
\footnotesize
\setlength{\tabcolsep}{1.5pt}
\setlength{\medmuskip}{2mu}
\setlength{\thickmuskip}{3mu}
\renewcommand{\arraystretch}{1.1}
\begin{tabular*}{\linewidth}{@{\extracolsep{\fill}}lrrrrr@{}}
\toprule
\textbf{Method} & \textbf{RMSE} & \textbf{DTWD} & \textbf{FD} &
\shortstack[r]{\textbf{Spurious}\\\textbf{attractors (\%)}} &
\shortstack[r]{\textbf{Distance}\\\textbf{from goals}} \\
\midrule
\multicolumn{6}{c}{\textit{2 attractors}} \\
\midrule
StableNODE ($\alpha{=}\epsilon{=}10^{-2}$) & $17.65 \pm 0.64$ & $8.37 \pm 1.35$ & $32.91 \pm 4.10$ & $0.31 \pm 0.63$ & {\boldmath$0.38 \pm 0.21$} \\
Single-attractor CONDORs + state machine & $77.85 \pm 3.94$ & $81.32 \pm 2.96$ & $106.28 \pm 9.11$ & $9.22 \pm 9.23$ & $3.71 \pm 0.70$ \\
Multi-CONDOR (only imitation, $\lambda_{\text{stable}} = \lambda_{\text{map}} = 0$) & {\boldmath$15.25 \pm 6.53$} & {\boldmath$5.69 \pm 1.48$} & {\boldmath$20.68 \pm 8.03$} & $16.98 \pm 20.89$ & $13.33 \pm 16.38$ \\
Multi-CONDOR (no mapping loss, $\lambda_{\text{map}}=0$) & $32.00 \pm 32.89$ & $23.62 \pm 42.24$ & $48.20 \pm 69.99$ & $13.33 \pm 35.19$ & $21.25 \pm 48.09$ \\
Multi-CONDOR (Ours) & $21.58 \pm 11.30$ & $8.82 \pm 6.97$ & $24.26 \pm 16.15$ & {\boldmath$0.00 \pm 0.00$} & $3.10 \pm 0.97$ \\
\midrule
\multicolumn{6}{c}{\textit{3 attractors}} \\
\midrule
StableNODE ($\alpha{=}\epsilon{=}10^{-2}$) & $13.46 \pm 0.51$ & $8.33 \pm 1.04$ & $24.50 \pm 1.48$ & $7.29 \pm 6.09$ & $17.90 \pm 10.68$ \\
Single-attractor CONDORs + state machine & $49.70 \pm 1.86$ & $60.03 \pm 1.96$ & $88.22 \pm 3.47$ & $13.75 \pm 6.63$ & $3.76 \pm 0.72$ \\
Multi-CONDOR (only imitation, $\lambda_{\text{stable}} = \lambda_{\text{map}} = 0$) & {\boldmath$11.26 \pm 7.09$} & {\boldmath$6.58 \pm 4.87$} & {\boldmath$18.49 \pm 14.57$} & $37.13 \pm 21.20$ & $28.24 \pm 30.37$ \\
Multi-CONDOR (no mapping loss, $\lambda_{\text{map}}=0$) & $13.23 \pm 10.27$ & $8.85 \pm 8.85$ & $20.15 \pm 18.58$ & $7.38 \pm 13.56$ & $3.77 \pm 3.14$ \\
Multi-CONDOR (Ours) & $14.08 \pm 10.54$ & $9.24 \pm 8.01$ & $21.18 \pm 17.10$ & {\boldmath$0.00 \pm 0.00$} & {\boldmath$2.69 \pm 0.79$} \\
\bottomrule
\end{tabular*}
\vspace{-5pt}
\end{table*}